\documentclass[letterpaper, 10pt, conference]{ieeeconf}

\IEEEoverridecommandlockouts                              

\usepackage{color}
\usepackage{graphics} 
\usepackage{epsfig} 
\usepackage{mathptmx} 
\usepackage{times} 
\usepackage{amsmath} 
\usepackage{amssymb}  

\title{\LARGE \bf
Coping with the variability in humans’ reward during simulated human-robot interactions through the coordination of multiple learning strategies* }

\author{R{\'e}mi Dromnelle$^{1}$, Beno{\^i}t Girard$^{1}$, Erwan Renaudo$^{2}$, Raja Chatila$^{1}$, Mehdi Khamassi$^{1}$ %
\thanks{* This work has been funded by a DGA (French National Defence Armaments Procurement Agency) scholarships (RD) and by Labex SMART (ANR–11– LABX–65).
The research leading to these results has received funding from the European Union’s Horizon 2020 research and innovation programme (grant agreement number 731761, IMAGINE).}%
\thanks{$^{1}$R{\'e}mi Dromnelle, Beno{\^i}t Girard, Raja Chatila and Mehdi Khamassi are with the Institute of Intelligent Systems and Robotics, Sorbonne Universit{\'e}, Paris, France}%
\thanks{$^{2}$Erwan Renaudo is with the Intelligent and Interactive Systems Group, Universit{\"a}t Innsbruck, Innsbruck, Austria}%
\thanks{Corresponding author: {\tt\small remi.dromnelle@isir.upmc.fr}}%
}

\begin{document}

\maketitle
\thispagestyle{empty}
\pagestyle{empty}

\begin{abstract}
An important current challenge in Human-Robot Interaction (HRI) is to enable robots to learn on-the-fly from human feedback. However, humans show a great variability in the way they reward robots. 
We propose to address this issue by enabling the robot to combine different learning strategies, namely model-based (MB) and model-free (MF) reinforcement learning. We simulate two HRI scenarios: a simple task where the human congratulates the robot for putting the right cubes in the right boxes, and a more complicated version of this task where cubes have to be placed in a specific order. We show that our existing MB-MF coordination algorithm previously tested in robot navigation works well here without retuning parameters. It leads to the maximal performance while producing the same minimal computational cost as MF alone. Moreover, the algorithm gives a robust performance no matter the variability of the simulated human feedback, while each strategy alone is impacted by this variability. Overall, the results suggest a promising way to promote robot learning flexibility when facing variable human feedback.
\end{abstract}

\section{INTRODUCTION}
Deploying autonomous robots in human societies requires these robots to be able to cope with the variability of feedback provided by humans. In particular, when human users want to teach a robot a new task, the robot cannot be retuned or reconfigured specifically for each human. The robot needs to be able to learn, no matter if the human user with which it interacts is generous or stingy in terms of reward, no matter if the human user consistently rewards the same actions or rather only initially rewards correct actions and then stops to do so. Previous HRI studies involving robot learning from human feedback have indeed highlighted such important variability in the way humans interact with the robots \cite{Najar2017}.

While this problem is rather novel and challenging for the field of HRI, it is striking that human children easily cope with the variability of their parents'/teachers' feedback \cite{ho2015}. Some parents tend to reward their children for every correct intermediate actions (\textit{e.g.,} for correctly placing each lego brick until finalizing a tower). Some reward only the first actions and later on consider remain silent while considering that the correctness of these actions remains true and implicit. Finally, some parents rarely reward correct action but rather limit their feedback to signaling incorrect actions \cite{roe1963}.

Whereas shaping reward relative to the average feedback of a context (\textit{e.g.,} related to a specific human user) could be a solution to transform the absence of punishments for some actions as rewarding \cite{palminteri2015}, this would not permit to cope with the non-stationarity of human users who initially reward correct actions and then stop to do so.

Past HRI studies show that increasing environmental rewards by humans can accelerate learning and reduce unwanted behaviours \cite{knox2010,griffith2013}. Mostly, in interactive learning tasks, robot teaching is supported by model-free (MF) reinforcement learning \cite{knox2010,Najar2017,qureshi2017}, and more rarely by model-based (MB) reinforcement learning \cite{fournier2019}. Usually, as long as humans train the robot well, MF learning is sufficient, but if humans give unreliable feedback, MF learning does not converge well. In contrast, MB learning converges more quickly and more robustly to the solution than MF learning, but is also computationnally more expensive \cite{SuttonB1998}. Here, we propose to equip social robots with the ability to coordinate on-the-fly the two approaches: a slow-learning but computationally cheap decision-making one (MF reinforcement learning) and a fast-learning but computationally expensive decision-making one (MB reinforcement learning). We test this idea here by simulating a recent coordination algorithm that we previously successfully apply to robot navigation in non-stationary environments \cite{dromnelleLivingMachines2020}. 

In this work, a metacontroller decides at every time step whether the MB or the MF strategy shall control the robot's actions, depending on which one has the highest learning quality while minimizing decision-making (inference) computational cost.

We test the algorithm in two simulated HRI tasks of increasing complexity, where the robot can autonomously learn to solve the tasks, and where a human can help it by providing social signals. In the first one, the robot learns to put a red cube in a red box, a green cube in a green box and a blue cube in a blue box. We simulate various humans with various rewarding tendencies and show that the MF strategy alone learns very slowly but can be bootstraped by human feedback; that the MB strategy alone learns very fast whether there is human feedback or not, but has a high computational cost; our MBMF coordination algorithm produces the optimal performance while minimizing the cost by switching to MF control of robot action once learning has converged. In the second task, cubes shall be placed in the boxes in a specific order, otherwise the task is failed. In this case, even the MB strategy alone is slower to learn and can be boostraped by human feedback. Once again, the MBMF coordination algorithm leads to optimal performance while minimizing computational cost. Overall, the proposed algorithm enables the robot's performance to be less impacted by the variability of human feedback.

\section{Materials and Methods}

\subsection{A robotic architecture with a dual decision-making system}

The present work implements a classical three-layer robot cognitive architecture \cite{Gat1998,AlamiCFGI1998} composed of a decision, an executive and a functional layer.

\begin{figure}[thpb]
\centering
\framebox{\includegraphics[scale=0.31]{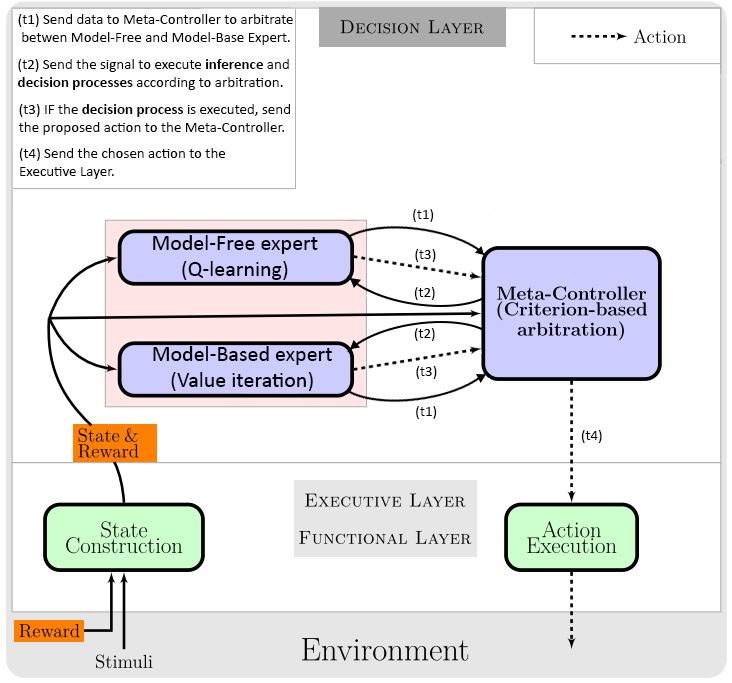}}
\caption{The generic version of the architecture. Two experts having different properties are computing the next action to do in the current state $s$. They each send monitoring data to the meta-controller (MC) about their learning status and inference process (t1). The MC chooses an expert according to a criterion that uses this data and authorizes it to carry out his inference and decision processes (t2). After the decision, the chosen expert sends its proposition to the MC (t3), which sends the action to the Executive Layer (t4). The effect of the executed action generates a new perception, transformed into an abstract Markovian state, and eventually a non null reward $r$, that are sent back to the experts. Each expert learns according to the action chosen by the MC, the new state reached and the reward.}
\label{fig:archi}
\end{figure}

The decision layer of the proposed architecture (Fig. \ref{fig:archi}) is composed by two competing experts which generate action propositions, each with its own method and with its own advantages and disadvantages. These two experts are directly inspired by the currently conventional distinction in computational neuroscience models between goal-directed and habitual strategies \cite{DawND2005}, the former being employed by humans at the beginning of learning, and requiring planning, while the latter is a form of automatization which enable fast decisions after learning has converged. The two experts run three processes in a row: learning, inference and decision. This layer is also provided with a meta-controller (MC) in charge of arbitrating between experts. The MC determines which expert's proposed action will be executed in the current state, according to an arbitration criterion, described below.

After that, the decision layer sends the chosen action to the executive layer, who ensures its accomplishment by recruiting robot's skills from the functional layer. The latter consists of a set of reactive sensorimotor loops that control actuators during interaction with the environment. The robot reaches a new state and obtains or not a reward. The two experts use the new state and the reward information to update their knowledge about the executed action. This allows MB and MF experts to cooperate by learning from each others decision. 

\subsection{The decision layer}

\subsubsection{Model-based (MB) expert}

The MB expert is implemented as a MB algorithm that learns a model ($\hat{T},\hat{R}$) of the transition function $T$ and the reward function $R$ of the problem, and uses this model to compute the values of actions in each state. This model allows the robot to simulate on several steps the consequences of following a given behavior and to quickly look for the desirable states to reach. However, this search process is costly in terms of computation time as it needs to simulate several iterations of value iteration \cite{SuttonB1998} in each state to find the correct solution.

\paragraph{Learning process}

For the MB expert, performing the learning process consists in updating its model ($\hat{T},\hat{R}$) by interacting with the world. The modeled transition function $\hat{T}$ is learnt by counting occurrences of transitions $(s,a,s')$:

$$
\hat{T}(s,a,s') = \frac{V_{N}(s,a,s')}{V_{N}(s,a)}
\eqno{(1)}
$$

\noindent where $V_{N}(s,a)$ is the number of visits of state $s$ and action $a$ (with a maximum value of $N$) and $V_{N}(s,a,s')$ is the number of visits of the transition $(s,a,s')$ in the last $N$ visits of $(s,a)$. This leads to an estimation of the probability to the closest multiple of $1/N$.

The modeled reward function $\hat{R}$ stores the most recent reward value $r_t$ received for performing action $a$ in state $s$ and reaching the current state $s'$, multiplied by the probability of the transition (s,a,s').

\paragraph{Inference process}

Performing the inference process consists in planning using a tabular Value Iteration algorithm \cite{SuttonB1998}: 

$$
Q(s,a) \leftarrow \sum_{s'} \hat{T}(s,a,s') \left[ \hat{R}(s,a) + \gamma U(s') \right]
\eqno{(2)}
$$

\noindent where $Q(s,a)$ is the action-value estimated by the agent for performing the action $a$ in the state $s$, $\gamma$ is the decay rate of future rewards and $U(s')$ is the maximum action-value of the state $s'$ estimated by the agent.

\paragraph{Decision process}

Performing the decision process consists in converting the estimation of action-values into a distribution of action probabilities using a softmax function, and drawing the action proposal from this distribution:

$$
P(a | s) = \frac{ \exp( Q(s,a) / \tau )  }{ \sum_{b \in \mathcal{A}} \exp(Q(s,b) / \tau )}
\eqno{(3)}
$$

where $\tau$ is the exploration/exploitation trade-off parameter.

\subsubsection{Model-free (MF) expert}

The MF algorithm does not use models of the problem to decide which action to do in each state, but directly compares action values to make a quick decision. These action values $Q(s,a)$ are progressively learned through direct experience with the world, and slowly estimating the average amount of expected future reward which follows the execution of each state-action pair. Because updating the action-values is local to the visited state, the process is slow. On the other hand, this method is less expensive in terms of inference time.

\paragraph{Learning process}

Performing the learning process therefore consists in estimating the action-value $Q(s,a)$ using a tabular Q-learning algorithm:

$$
Q(s,a) = Q(s,a) + \alpha \left[ R(s) + \gamma U(s') - Q(s,a) \right]
\eqno{(4)}
$$

\noindent where $R(s)$ is the scalar reward received for reaching the state $s$, $\gamma$ is the decay rate of future rewards and $U(s')$ is the maximum action-value of the state $s'$ estimated by the agent.

\paragraph{Inference process}

Since the MF expert does not use planning, its inference process consists only in reading from the table that contains all the action-values the one that corresponds to performing the action $a$ in the state $s$.

\paragraph{Decision process}

The decision process is the same as the one from the MB expert (Eq. 3). 

\subsubsection{\label{mcorga}Meta-controller (MC) and arbitration method}

The MC is in charge of selecting which expert will generate the behavior at each timestep. For each state $s$, it computes the entropy $H(s,E)$ of the action probability distribution of expert $E$, which can be used as a measure of the quality of learning:

$$
H(s,E) = - \sum_{a=0}^{|\mathcal{A}|} P(a|s)\cdot \log_{2}(P(a|s))
\eqno{(5)}
$$

\noindent where $P(a|s)$ is the probability of selecting action $a$ in state $s$. The lower the entropy, the lower the uncertainty of the agent about the action to choose. So the lower the entropy, the higher the quality of learning. The action selection probabilities used to compute the entropy are averaged over time per state using an exponential moving average.

For each state, the MC also computes the exponential moving average of the time taken to perform the inference process $T_{s,E}$ of expert $E$. The arbitration criterion we have proposed, and which was have previously applied to robot navigation, is a trade-off between the quality of learning and the cost of inference \cite{dromnelleLivingMachines2020}. By using it, the MC can deal between favouring the most certain expert (the most efficient) and the cheapest expert in terms of calculation. To do this, it computes one expert-value $Q(s,E)$ (6) for each expert.

$$
Q(s,E) = - (H(s,E) + \kappa T(s,E))
\eqno{(6)}
$$

$$
\kappa = e^{-9 H(MF)}
\eqno{(7)}
$$

\noindent where $\kappa$ is a parameter fixed by the experimenter so as to weight the impact of time: the lower the entropy of the distribution of action probabilities, the more weight the time taken to perform the inference process has in the equation. The action selection probabilities used to compute the entropy are averaged over time per state using an exponential moving average. We have chosen the value of $\kappa$ according to a Pareto front analysis \cite{powel2015} (not shown here). We were looking for a $\kappa$ that maximizes the quality of learning, minimizes the cost of inference, while maximizing the agent's ability to accumulate reward over time. 

Finally, the MC converts the estimation of expert-values $Q(s,E)$ into a distribution of expert probabilities using a softmax function, and drawing an expert from this distribution. The inference process of the unchosen expert is inhibited, which thus allows the system to save computing time.

\subsubsection{General information}
For this study, we reuse the parameters from our previous navigation study \cite{dromnelleLivingMachines2020}, except for the parameter $\kappa$ of the MC. For the MF expert, $\alpha$ = 0.6 and $\gamma$ = 0.9. For the MB expert, $\gamma$ = 0.95. For the MF, the MB and the MC expert, $\tau$ = 0.02. We initialized all the action-values to a null value.

\subsection{The social interaction task}

\subsubsection{The tidy task}

The virtual robot is a mechanical arm with visual sensors and its environment is a table on which three cubes and three boxes of red, green, and blue colors are placed. Since our coordination model is composed of tabular learning algorithms, the environment is represented by a discrete state space.

In the first version of the task (Fig.~\ref{fig:task}.A), a state is defined by the position of each of the three cubes: either on the table, in the red box, in the green box, in the blue box or in the robot hand. Note that only one cube at a time can be held by the hand. This represents a total of 112 different states. In this task, the objective of the robot is to learn how to put each cube in the box of the corresponding colour. When it is done, the robot obtains a virtual reward unit, and the cubes are automatically placed back on the table for the next trial.

In the second version of the task (Fig.~\ref{fig:task}.B), a notion of order is added. The objective of the robot is to learn to first put the red cube in the red box, then the green cube, then the blue one, without interrupting the sequence (any failure requires to restart the sequence from the beginning). To model this, a three-LED lamp is added to the simulated environment, indicating the validity of the sequence, and its state is added to the state description used in learning. When the robot puts the red cube in the red box, the first LED lights up, when it puts the green cube in the green box after having put the red cube in the red box, the second LED lights up, etc. If the robot make an error during the execution of the sequence, all the LEDs will turn off. This second version of the task is modeled by a set of 147 states. When all LEDs are on, the robot obtains a virtual reward unit, as for the first task.

When the robot reaches the final state, all the cubes are replaced on the table. To interact with the environment, the robot can perform a set of 7 different actions: take the red cube, take the green cube, take the blue cube, place the hand-held cube in the red box, place the hand-held cube in the green box, place the hand-held cube in the blue box and place the hand-held cube on the table. While other ways of modeling the task, such as with relational reinforcement learning \cite{dvzeroski2001}, would have been possible, we chose this state and action decomposition for simplicity as proof-of-concept of the strength of combining MB and MF learning strategies in HRI.

Finally, we have defined a long babbling period of 1000 iterations, where the robot explores its environment without reward and Human interactions, in order to allow the MB expert to build a workable transition model.

\subsubsection{Simulating social interactions}

A simulated human who can interact with the robot is facing the table. We have defined two ways for the robot to learn from humans.

\begin{figure}[thpb]
\centering
\framebox{\includegraphics[scale=0.08]{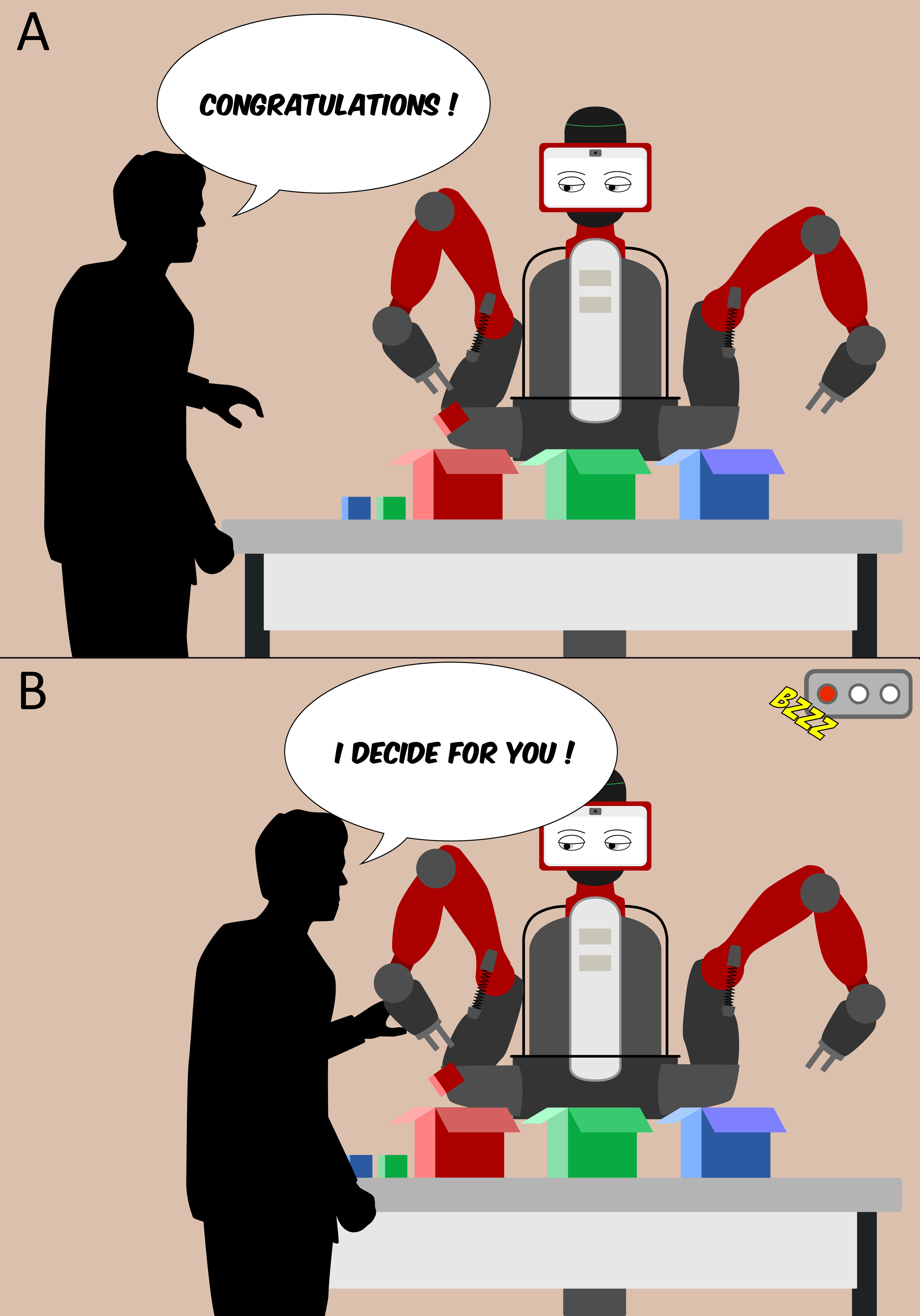}}
\caption{A. The first version of the task and the congratulation type interaction. B. The second version of the task and the takeover type interaction. The version of the task and the nature of the interaction are permutable. }
\label{fig:task}
\end{figure}

The human can congratulate the robot after it performs a relevant action (Fig.~\ref{fig:task}.A), for example when the robot puts the red cube in the red box for the first task, or when it puts the green cube in the green box after the red cube is put in the red box for the second task. The effect of the interaction will be effective the next time the robot will be in this situation. \cite{stone2012} have shown that the more directly human congratulation affects the robot's action selection process, the better it is, and the more it affects the updating of action-values for each transition experiment, the worse it is. 

Thus, in our work, we model congratulation as a bias of an action-value only during the decision process, and not as a direct modification of the action-values model. Therefore, the decision-making in Eq. 3 is now computed using the learned Q-value plus a one-time bonus provided if the human gave congratulations the last time the situation was met:

$$
P(a | s) = \frac{ \exp(( Q(s,a)+\zeta*{G(s,a)}) / \tau )  }{ \sum_{b \in \mathcal{A}} \exp((Q(s,b)+\zeta*{G(s,b)}) / \tau )}
\eqno{(8)}
$$

\noindent where $G(s,a)$ equals to 1 if the human guide congratulated the $(s,a)$ combination the last time it was executed, and 0 otherwise. We set the $\zeta$ parameter to 0.1. 

The human can also override the robot's choice, when a cube is held by the robot, by choosing where it will be dropped (Fig.~\ref{fig:task}.B). Here, the interaction has an instant effect on the robot. It can't be act, but still learns from the observation of the consequences of this Human-chosen actions.

To take into account the fact that a human is not always perfectly rigorous, we considered that he forgets to interact with the robot about 10\% of the time when he faces the robot.

\section{RESULTS}

\begin{figure}[t]
\centering
\framebox{\includegraphics[scale=0.090]{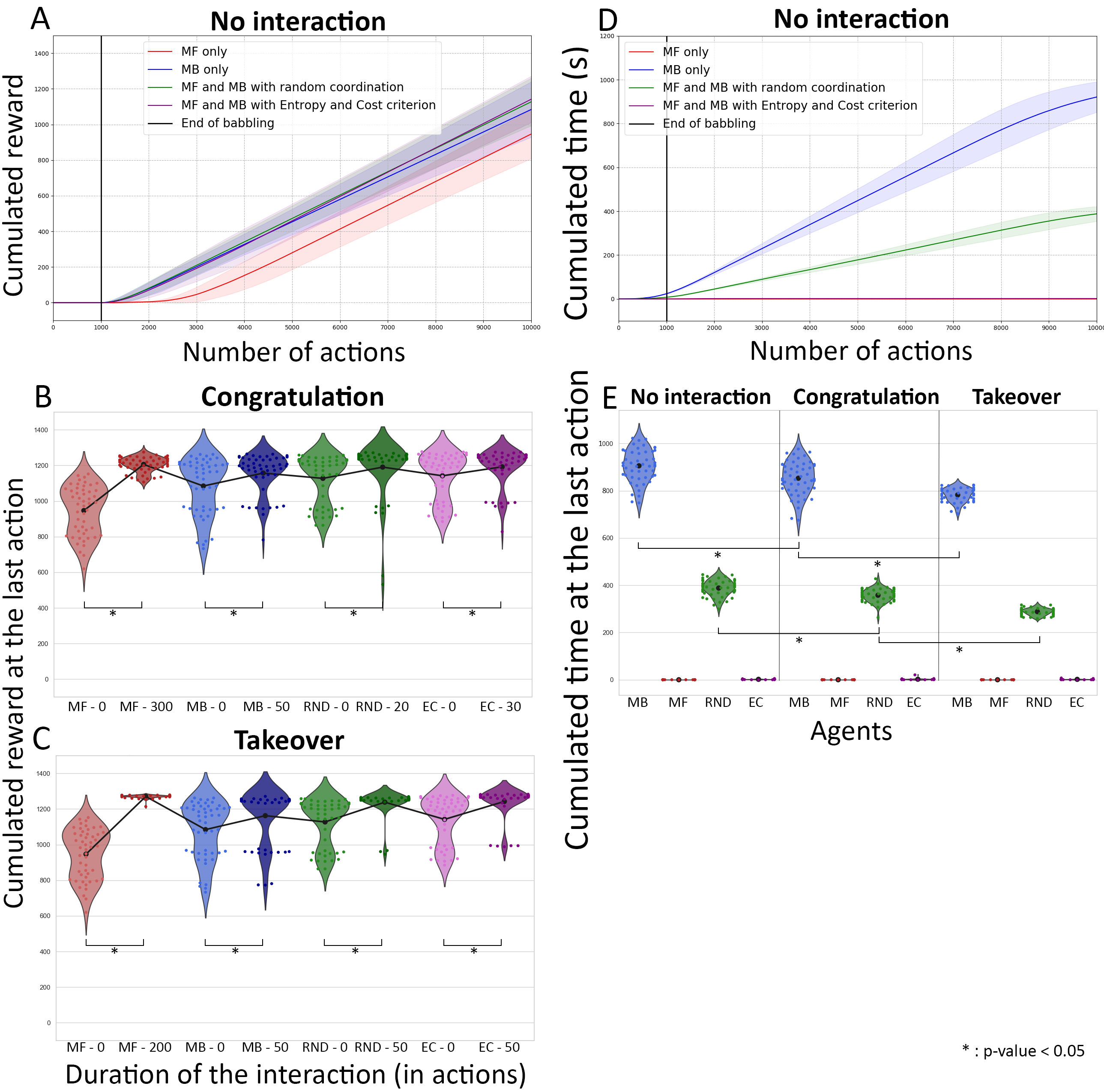}}
\caption{Task 1 results. A. Reward accumulation without Human interactions. B. Cumulated reward after 10k actions, for the MF (red), MF (blue), RND (green) and EC (purple) robots, with no interactions (light) or optimal number of Congratulation interactions (dark). C. Same for Takeover interactions. D. Computation cost accumulation without interactions. E. Cumulated computation time for the different robots (same color code as B), without interactions (left), with congratulations (middle) or with takeover (right) after 10k actions. Statistical significance in the difference assessed with Wilcoxon-Mann-Whitney tests.}
\label{fig:results1}
\end{figure}

\begin{figure}[thpb]
\centering
\framebox{\includegraphics[scale=0.127]{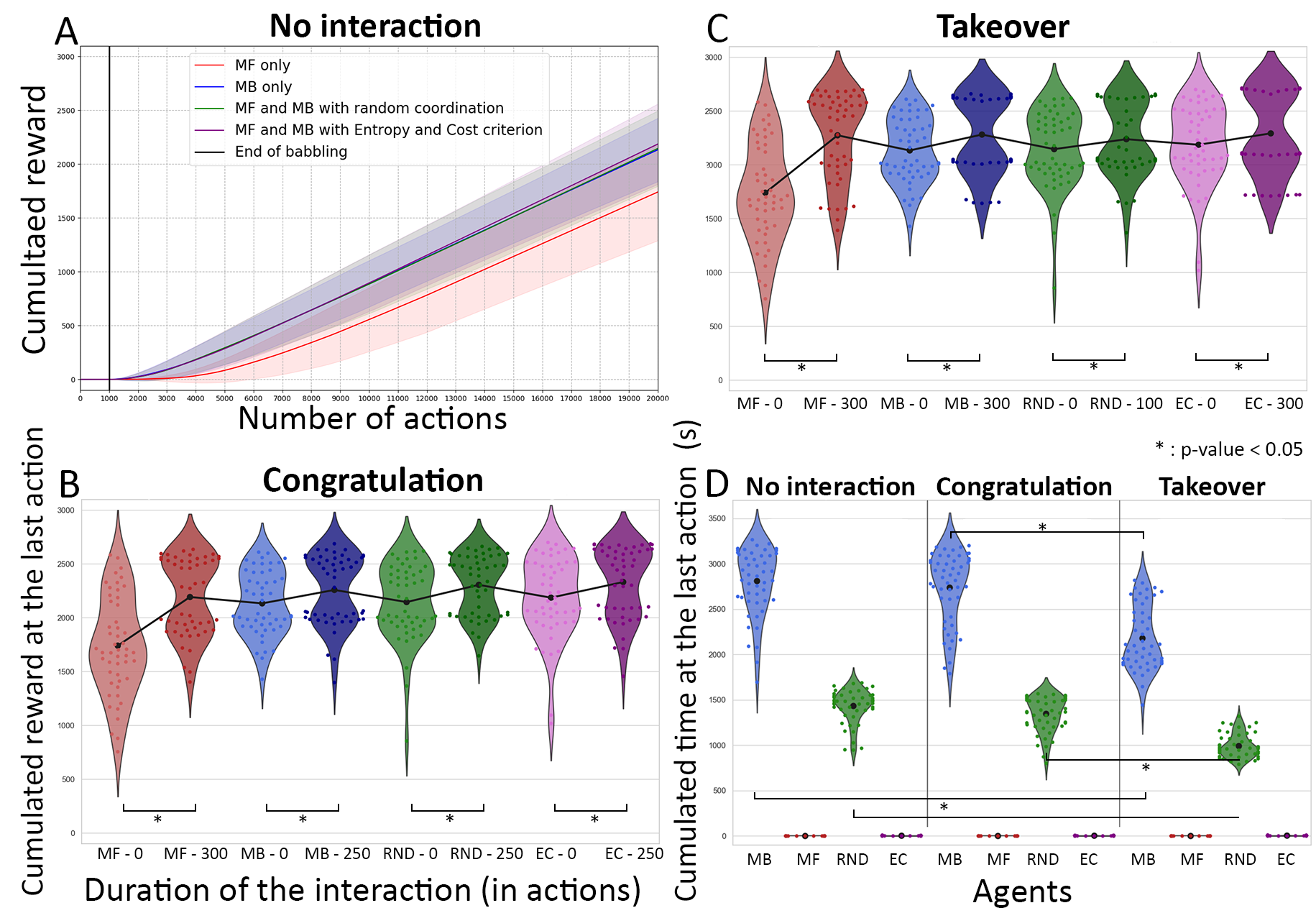}}
\caption{Task 2 results. A. Reward accumulation without Human interactions. B. Cumulated reward after 20k actions, for the different robots, with no interactions or optimal number of Congratulation interactions. C. Same for Takeover interactions. D. Cumulated computation time for the different robots, without interactions, with congratulations or with takeover after 20k actions. Statistical significance in the difference assessed with Wilcoxon-Mann-Whitney tests.}
\label{fig:results3}
\end{figure}

To evaluate the performance of the virtual robot, we studied four combinations of experts: (1) an MF expert alone (in red in Figs.~\ref{fig:results1} and \ref{fig:results3}), (2) an MB expert alone (in blue), (3) an MC choosing randomly at each timestep between MF and MB (RND, in green), and (4) an MC using the Entropy and Cost arbitration criterion presented in \ref{mcorga} (EC, in purple). For each combinations, 50 runs were simulated. 

Fig. \ref{fig:results1}.A represents the evolution of the reward accumulated by robots over time, in the first task, without Human interactions. In this situation, the MF robot needs more time to start to accumulate reward at steady rate. On the other hand, while the MB robot starts steadily accumulating reward earlier, he also converges to an accumulation rate lower than the MF's one, so that in the long run, it will be caught up by the MF robot. The RND and EC robots seem to combine the best of both worlds: fast convergence to the maximal accumulation rate.

For each robot controller, we have experimentally determined (data not shown) the minimal number of Human interactions necessary to reach the maximal possible accumulation rate (i.e. more interactions do not improve it): 300 for MF, 50 for MB, 20 for RND and 30 for EC, for the congratulation type of interactions, and 200 for MF and 50 for MB, RND, and EC for the takeover type of interactions.
Fig. \ref{fig:results1}.B compares the performance of the different robots with or without the congratulation interaction: with congratulations, the MF robot shows a higher performance score than the MB robot, but requires more interactions to reach it (300 vs. 50). The robots (RND and EC) which coordinate MF and MB experts, exhibit a performance equivalent to the one of the MF robot, but with much less interactions (20 and 30, vs. 300). With the takeover type of interactions the results follow the same trends, but are stronger (Fig. \ref{fig:results1}.C): in all cases, the interactions have a significant effect on the performance improvement, that this effect is the strongest for the MF robot, and the final levels are higher than with congratulations.

Concerning the computational cost of the different robot controllers without Human interactions, as expected, the MB robot is much more costly than the MF (see the accumulation of computation time on 
Fig. \ref{fig:results1}.D). The RND, which statistically uses each of them half of the time, has a time cost in between. Finally, the EC is much more efficient, as its time cost is very close to the one of the MF.

The addition of Human interactions also improves the time cost of MF and RND (Fig. \ref{fig:results1}.E), again with stronger effects for the takeover type of interactions.
These results show the interest to use a the EC coordination system instead of the purely random one, which exhibits the same reward accumulation performance, but which is much more expensive.

\begin{figure}[t]
\centering
\framebox{\includegraphics[scale=0.133]{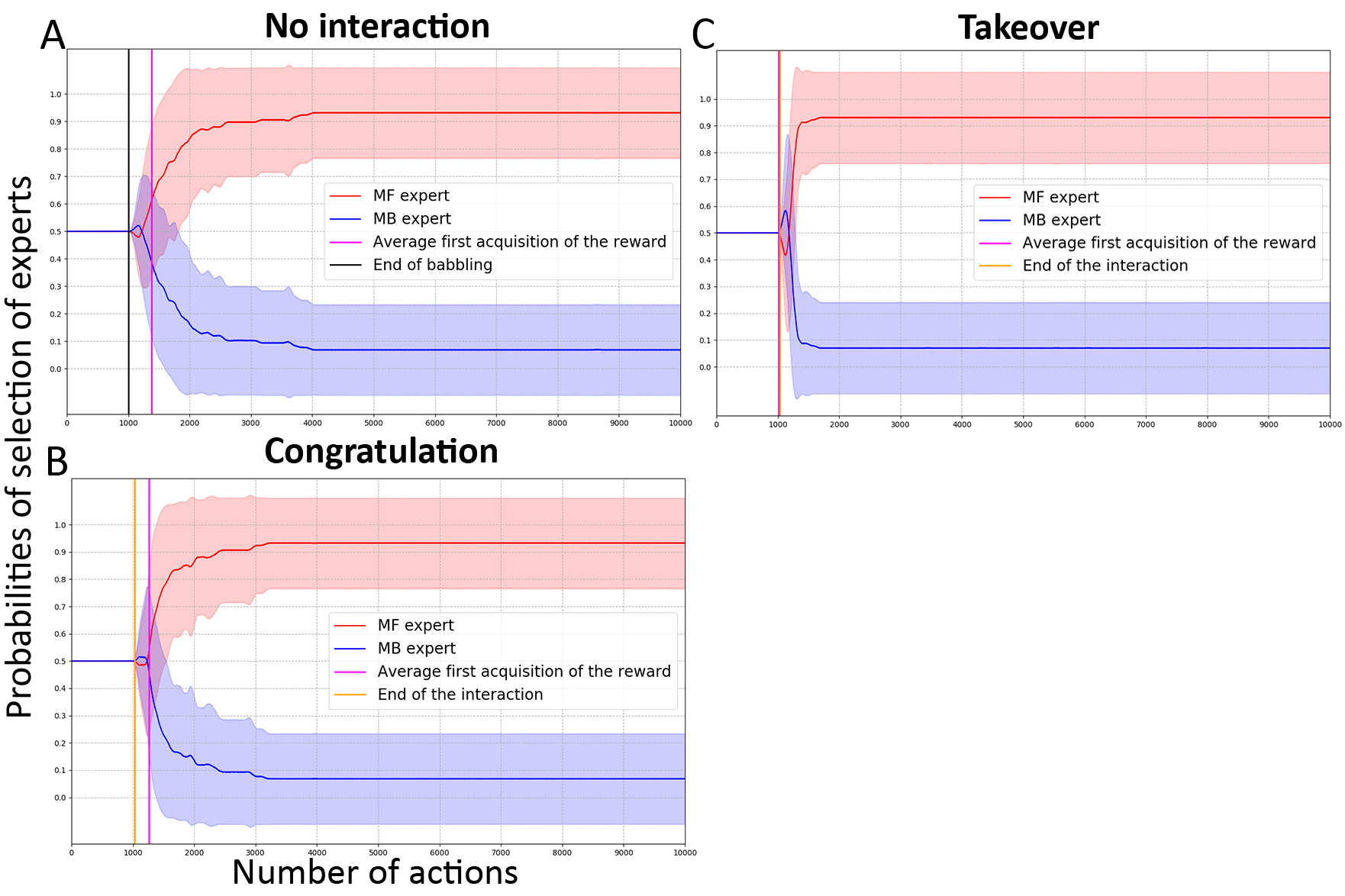}}
\caption{Task 1 results. A. Evolution of the probabilities of selection of the MF (red) and the MB (blue) expert by the meta-controller of the EC robot, without Human interactions. B. Same for Congratulation interactions. C. Same for Takeover interactions.}
\label{fig:results2}
\end{figure}

Fig. \ref{fig:results2}.A shows the dynamics of the selection of the experts by the meta-controller for the EC robot, expressed in terms of selection probabilities. Directly after the end of the babbling phase, the MB expert has on average more the control on the decision than the MF expert, which explains why the EC robot exhibits the same performance as the MB robot at the beginning of the experiment. Later on, the MF expert largely takes the control over the decision for the rest of the experiment, which greatly lowers the computational cost of the system. We observe the same temporal pattern with the two types of Human interactions (Fig. \ref{fig:results2}.B and Fig. \ref{fig:results2}.C). Also, with Human interactions, and especially the takeover one, the first reward is on average obtained earlier than without interactions.

To sum up, we saw that without Human interactions, using the MB robot seems to be a good alternative in terms of performance, but is expensive in terms of computing resources. With Human interactions, using the MB expert helps the robot a little bit, but the effect of the interactions is not huge. For the MF robot, Without Human interactions, it takes a lot of time to learn how to perform the task, and on the other hand, it costs almost nothing in terms of computational resources. With Human interactions, the MF robot becomes more accurate than the approximate model learned by the the MB expert, even if it requires a huge number of interactions. Interestingly, we showed that with or without Human interactions, the EC robot reaches the maximal accumulation rate of reward while costing as little as the MF robot in terms of computing resources. So, finally, using a wise coordination system such as the EC robot allows the robot to be more autonomous by avoiding the human dependency at a lower cost. Such a system is useful when human participation during the task is uncertain or not very consistent, \textit{i.e.,} the human initially rewards the robot and later on stops to do so, as we have simulated here.

We replicate these conclusions with the second more complicated task (Fig. \ref{fig:results3}), where robots take longer to accumulate reward over time. After 10.000 iterations, the MF robot shows a performance loss of about 40\%, of about 20\% for the MB robot and of about 30\% for the RND and the EC robots, in comparison to the first task. To give the MF robot time to learn effectively how to solve the task, we have extended the simulations to 20.000 iterations. Globally, the same effects of the interactions on the performance of the robots are observed (Fig. \ref{fig:results3}.B and Fig. \ref{fig:results3}.C). The main difference, and a not negligible one, is the much higher number of interactions needed to obtain the maximum effect on robots' performance. For the MF robot, this does not change anything, and about 300 actions are still necessary. On the other hand, for the other robots, while a maximum of 50 actions were sufficient in the first task, several hundred are needed here. The task is therefore definitely more complicated. Even so, the EC robot shows again the best performance for a derisory computational cost (Fig. \ref{fig:results3}.D), with or without Human interactions. 

\section{CONCLUSION}

Mostly, robot teaching is supported by model-free (MF) reinforcement learning. As long as humans train the robot in the right way and consistently, it is sufficient. Unfortunately, this is not necessarily the case all the time. The alternative could be to use a model-based (MB) reinforcement learning, which is generally more efficient and especially much less dependent on human guidance. Unfortunately, it is also much more expensive in terms of computing resources. In this work, we show the interest to use in HRI an hybrid MB-MF algorithm that takes advantage of each of the strategies. To illustrate this, we use the coordination algorithm presented in \cite{dromnelleLivingMachines2020}, and initially applied to a non-social navigation task, without significant retuning of the parameters. 

We show that our hybrid system leads to the maximal performance while producing the same minimal computational cost as MF alone. Moreover, it gives a robust performance no matter the variability of the simulated human feedback, while each strategy alone is impacted by this variabiliy, and requires less interactions with the human to converge to a stable performance than the MB or the MF alone. We also show that doing the action instead of the robot appears to be more efficient than a posterior congratulation, even if it means a higher involvement of the human.
In conclusion, we highlight that the judicious use of a model-based expert, in support of a model-free expert, allows the robot to accelerate its learning and also to maintain for free its level of performance when confronted with a human little involved in the task. Here, the human can be considered as an expert who knows the task better than others, and the MB expert as the rescue expert, who supports the MF expert when the human disengages. These results suggest a promising way to promote the flexibility of robot learning in the face of variable human reactions.

In futur work, we plan to test other types of interactions such as punishment or nuisance. It would also be interesting to study how the robot behaves if only humans can bring reward or if the objective of the task changes during the experiment \cite{dromnelleLivingMachines2020}. Finally, we want to reproduce this study with a real robot and real humans participants. 

\section*{ACKNOWLEDGMENT}
Thanks to Romain Retureau for the help on the illustrations of the experimental task.

\bibliographystyle{ieeetr}
\bibliography{article}

\end{document}